# Utilizing Statistical Dialogue Act Processing in Verbmobil

**Norbert Reithinger and Elisabeth Maier***
DFKI GmbH
Stuhlsatzenhausweg 3
D-66123 Saarbrücken
Germany
{reithinger,maier}@dfki.uni-sb.de

## Abstract

In this paper, we present a statistical approach for dialogue act processing in the dialogue component of the speech-to-speech translation system VERBMOBIL. Statistics in dialogue processing is used to predict follow-up dialogue acts. As an application example we show how it supports repair when unexpected dialogue states occur.

## 1 Introduction

Extracting and processing communicative intentions behind natural language utterances plays an important role in natural language systems (see e.g. (Cohen et al., 1990; Hinkelman and Spackman, 1994)). Within the speech-to-speech translation system VERBMOBIL (Wahlster, 1993; Kay et al., 1994), dialogue acts are used as the basis for the treatment of intentions in dialogues. The representation of intentions in the VERBMOBIL system serves two main purposes:

- Utilizing the dialogue act of an utterance as an important knowledge source for translation yields a faster and often qualitative better translation than a method that depends on surface expressions only. This is the case especially in the first application of VERBMOBIL, the on-demand translation of appointment scheduling dialogues.
- Another use of dialogue act processing in VERBMOBIL is the prediction of follow-up dialogue acts to narrow down the search space on the analysis side. For example, dialogue act predictions are employed to allow for dynamically adaptable language models in word recognition. Recent results (e.g. (Niedermair, 1992)) show a reduction of perplexity in the word recognizer between 19% and 60% when context dependent language models are used.

Dialogue act determination in VERBMOBIL is done in two ways, depending on the *system mode*: using *deep* or *shallow processing*. These two modes depend on the fact that VERBMOBIL is only translating *on demand*, i.e. when the user's knowledge of English is not sufficient to participate in a dialogue. If the user of VERBMOBIL needs translation, she presses a button thereby activating deep processing. In depth processing of an utterance takes place in maximally 50% of the dialogue contributions, namely when the owner speaks German only. Dialogue act extraction from a DRS-based semantic representation (Bos et al., 1994) is only possible in this mode and is the task of the semantic evaluation component of VERBMOBIL.

In the other processing mode the dialogue component tries to process the English passages of the dialogue by using a keyword spotter that tracks the ongoing dialogue superficially. Since the keyword spotter only works reliably for a vocabulary of some ten words, it has to be provided with keywords which typically occur in utterances of the same dialogue act type; for every utterance the dialogue component supplies the keyword spotter with a prediction of the most likely follow-up dialogue act and the situation-dependent keywords.

The dialogue component uses a combination of statistical and knowledge based approaches to process dialogue acts and to maintain and to provide contextual information for the other modules of VERBMOBIL (Maier and McGlashan, 1994). It includes a robust dialogue plan recognizing module, which uses repair techniques to treat unexpected dialogue steps. The information acquired during dialogue processing is stored in a dialogue memory. This contextual information is decomposed into the intentional structure, the referential structure, and the temporal structure which refers to the dates mentioned in the dialogue.

*This work was funded by the German Federal Ministry for Education, Research and Technology (BMBF) in the framework of the Verbmobil Project under Grant 01IV101K/1. The responsibility for the contents of this study lies with the authors. Thanks to Jan Alexandersson for valuable comments and suggestions on earlier drafts of this paper.

An overview of the dialogue component is given in (Alexandersson et al., 1995). In this paper main emphasis is on statistical dialogue act prediction in VERBMOBIL, with an evaluation of the method, and an example of the interaction between plan recognition and statistical dialogue act prediction.

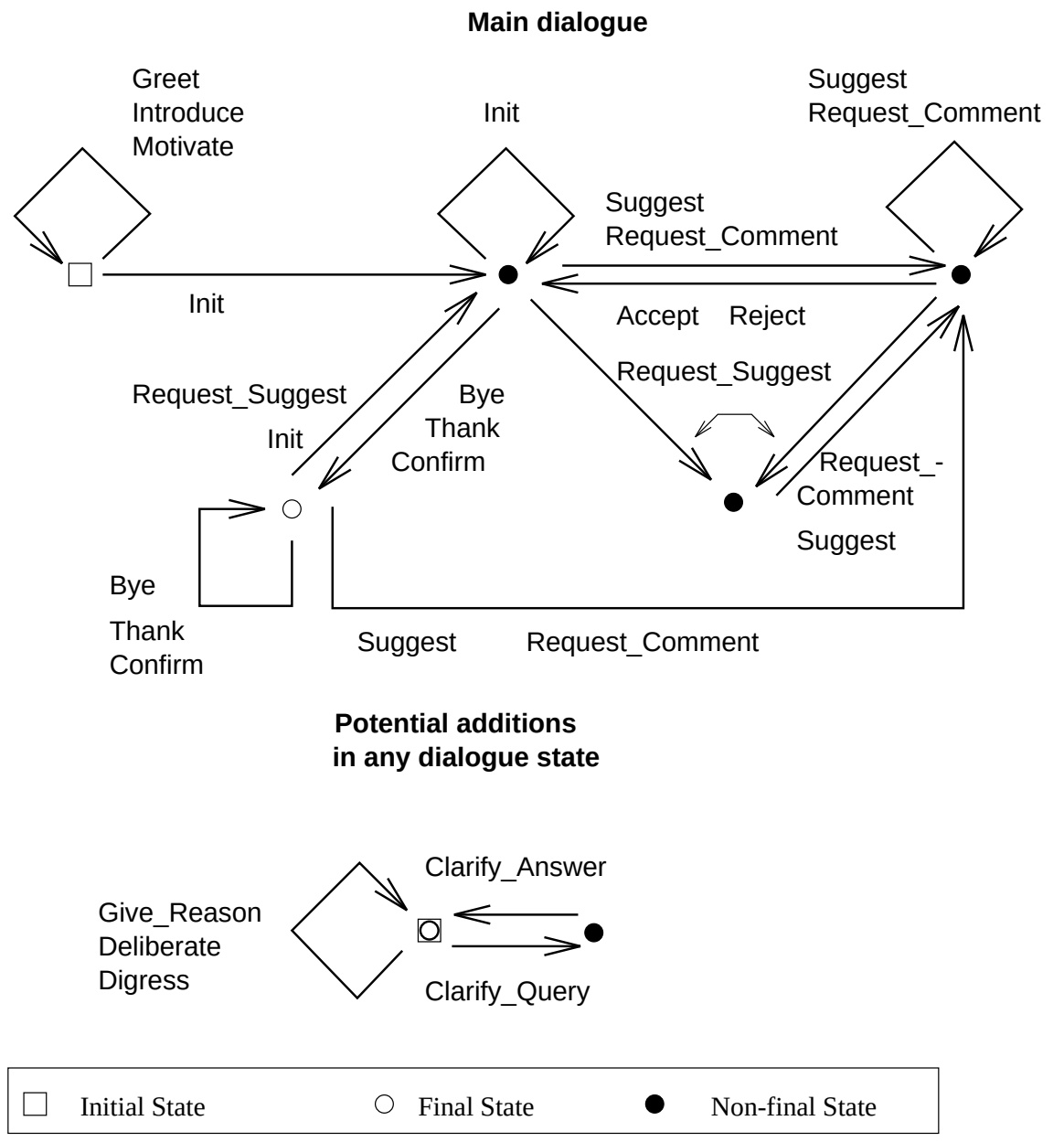

Figure 1: A dialogue model for the description of appointment scheduling dialogs

## 2 The Dialogue Model and Predictions of Dialogue Acts

Like previous approaches for modeling task-oriented dialogues we assume that a dialogue can be described by means of a limited but open set of dialogue acts (see e.g. (Bilange, 1991), (Mast et al., 1992)). We selected the dialogue acts by examining the VERBMOBIL corpus, which consists of transliterated spoken dialogues (German and English) for appointment scheduling. We examined this corpus for the occurrence of dialogue acts as proposed by e.g. (Austin, 1962; Searle, 1969) and for the necessity to introduce new, sometimes problem-oriented dialogue acts. We first defined 17 dialogue acts together with semi-formal rules for their assignment to utterances (Maier, 1994). After one year of experience with these acts, the users of dialogue acts in VERBMOBIL selected them as the domain independent "upper" concepts within a more elaborate hierarchy that becomes more and more propositional and domain dependent towards its leaves (Jekat et al., 1995). Such a hierarchy is useful e.g. for translation purposes. Following the assignment rules, which also served as starting point for the automatic determination of dialogue acts within the semantic evaluation component, we hand-annotated over 200 dialogues with dialogue act information to make this information available for training and test purposes.

Figure 1 shows the domain independent dialogue acts and the transition networks which define admissible sequences of dialogue acts. In addition to the dialogue acts in the main dialogue network, there are five dialogue acts, which we call deviations, that can occur at any point of the dialogue. They are represented in an additional subnetwork which is shown at the bottom of figure 1. The networks serve as the basis for the implementation of a parser which determines whether an incoming dialogue act is compatible with the dialogue model.

As mentioned in the introduction, it is not only important to extract the dialogue act of the current utterance, but also to predict possible follow up dialogue acts. Predictions about what comes next are needed internally in the dialogue component and externally by other components in VERBMOBIL. An example of the internal use, namely the treatment of unexpected input by the plan recognizer, is described in section 4. Outside the dialogue component dialogue act predictions are used e.g. by the abovementioned semantic evaluation component and the keyword spotter. The semantic evaluation component needs predictions when it determines the dialogue act of a new utterance to narrow down the set of possibilities. The keyword spotter can only detect a small number of keywords that are selected for each dialogue act from the VERBMOBIL corpus of annotated dialogues using the Keyword Classification Tree algorithm (Kuhn, 1993; Mast, 1995).

For the task of dialogue act prediction a knowledge source like the network model cannot be used since the average number of predictions in any state of the main network is five. This number increases when the five dialogue acts from the subnetwork which can occur everywhere are considered as well. In that case the average number of predictions goes up to 10. Because the prediction of 10 dialogue acts from a total number of 17 is not sufficiently restrictive and because the dialogue network does not represent preference information for the various dialogue acts we need a different model which is able to make reliable dialogue act predictions. Therefore we developed a statistical method which is described in detail in the next section.

## 3 The Statistical Prediction Method and its Evaluation

In order to compute weighted dialogue act predictions we evaluated two methods: The first method is to attribute probabilities to the arcs of our network by training it with annotated dialogues from our corpus. The second method adopted information theoretic methods from speech recognition. We implemented and tested both methods and currently favor the second one because it is insensitive to deviations from the dialogue structure as described by the dialogue model and generally yields better prediction rates. This second method and its evaluation will be described in detail in this section.

Currently, we use n-gram dialogue act probabilities to compute the most likely follow-up dialogue act. The method is adapted from speech recognition, where language models are commonly used to reduce the search space when determining a word that can match a part of the input signal (Jellinek, 1990). It was used for the task of dialogue act prediction by e.g. (Niedermair, 1992) and (Nagata and Morimoto, 1993). For our purpose, we consider a dialogue $S$ as a sequence of utterances $S_i$ where each utterance has a corresponding dialogue act $s_i$. If $P(S)$ is the statistical model of $S$, the probability can be approximated by the n-gram probabilities

$$P(S) = \prod_{i=1}^{n} P(s_i | s_{i-N+1}, ..., s_{i-1})$$

Therefore, to predict the $n$th dialogue act $s_n$ we can use the previously uttered dialogue acts and determine the most probable dialogue act by computing

$$s_n := \max_{s} \; P(s|s_{n-1}, s_{n-2}, s_{n-3}, ...)$$

To approximate the conditional probability $P(.|.)$ the standard smoothing technique known as deleted interpolation is used (Jellinek, 1990) with

$$P(s_n|s_{n-1}, s_{n-2}) =$$
$$q_1 f(s_n) + q_2 f(s_n|s_{n-1}) + q_3 f(s_n|s_{n-1}, s_{n-2})$$

where $f$ are the relative frequencies computed from a training corpus and $q_i$ weighting factors with $\sum q_i = 1$.

To evaluate the statistical model, we made various experiments. Figure 2 shows the results for three representative experiments (TS1-TS3, see also (Reithinger, 1995)).

| *Pred.* | *TS1* | *TS2* | *TS3* |
|---|---|---|---|
| 1 | 44,24 % | 37.47 % | 40.28 % |
| 2 | 66,47 % | 56.50 % | 59.62 % |
| 3 | 81,46 % | 69.52 % | 71.93 % |

Figure 2: Predictions and hit rates

In all experiments 41 German dialogues (with 2472 dialogue acts) from our corpus are used as *training* data, including deviations. TS1 and TS2 use the same 81 German dialogues as *test* data. The difference between the two experiments is that in TS1 only dialogue acts of the main dialogue network are processed during the test, i.e. the deviation acts of the test dialogues are not processed. As can be seen — and as could be expected — the prediction rate drops heavily when unforseeable deviations occur. TS3 shows the prediction rates, when all currently available annotated dialogues (with 7197 dialogue acts) from the corpus are processed, including deviations.

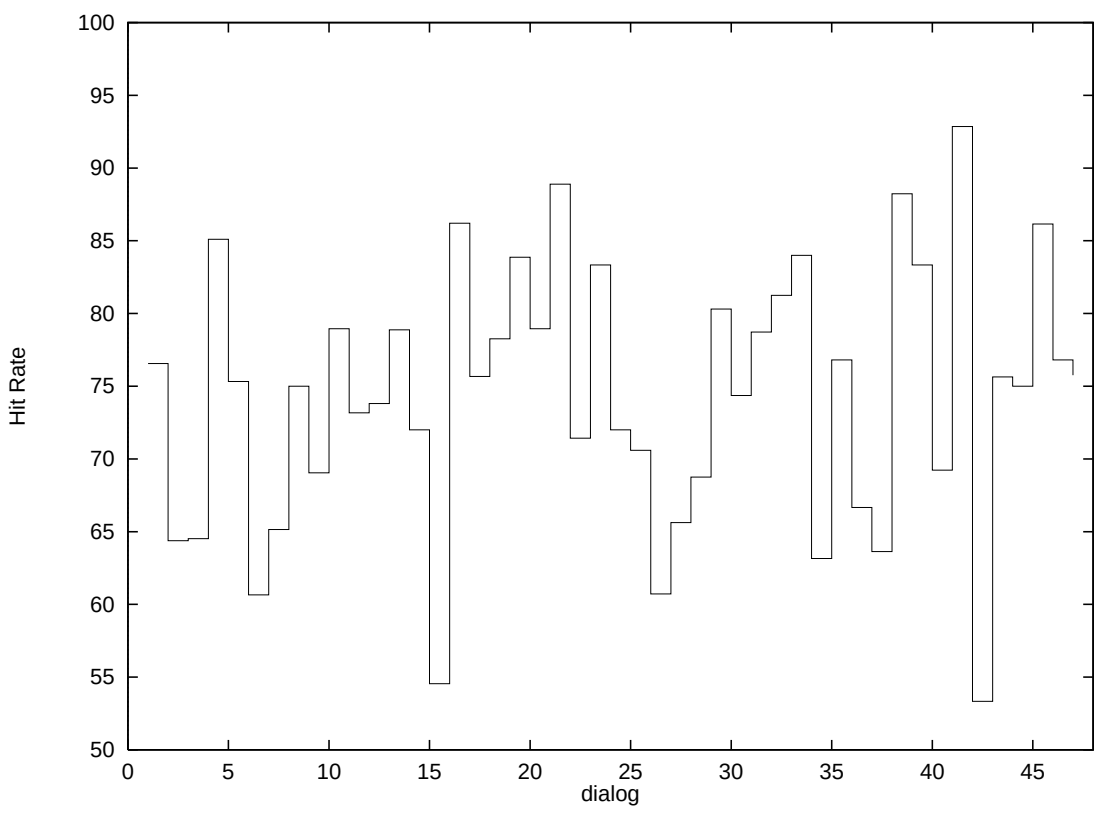

Figure 3: Hit rates for 47 dialogues using 3 predictions

Compared to the data from (Nagata and Morimoto, 1993) who report prediction rates of 61.7 %, 77.5 % and 85.1 % for one, two or three predictions respectively, the predictions are less reliable. However, their set of dialogue acts (or the equivalents, called illocutionary force types) does not include dialogue acts to handle deviations. Also, since the dialogues in our corpus are rather unrestricted, they have a big variation in their structure. Figure 3 shows the variation in prediction rates of three dialogue acts for 47 dialogues which were taken at random from our corpus. The x-axis represents the different dialogues, while the y-axis gives the hit rate for three predictions. Good examples for the differences in the dialogue structure are the dialogue pairs #15/#16 and #41/#42. The hit rate for dialogue #15 is about 54% while for #16 it is about 86%. Even more extreme is the second pair with hit rates of approximately 93% vs. 53%. While dialogue #41 fits very well in the statistical model acquired from the training-corpus, dialogue #42 does not. This figure gives a rather good impression of the wide variety of material the dialogue component has to cope with.

## 4 Application of the Statistical Model: Treatment of Unexpected Input

The dialogue model specified in the networks models all dialogue act sequences that can be usually expected in an appointment scheduling dialogue. In case unexpected input occurs repair techniques have to be provided to recover from such a state and to continue processing the dialogue in the best possible way. The treatment of these cases is the task of the dialogue plan recognizer of the dialogue component.

The plan recognizer uses a hierarchical depth-first left-to-right technique for dialogue act processing (Vilain, 1990). Plan operators have been used to encode both the dialogue model and methods for recovery from erroneous dialogue states. Each plan operator represents a specific `goal` which it is able to fulfill in case specific `constraints` hold. These constraints mostly address the context, but they can also be used to check pragmatic features, like e.g. whether the dialogue participants know each other. Also, every plan operator can trigger follow-up `actions`. A typical action is, for example, the update of the dialogue memory. To be able to fulfill a goal a plan operator can define `subgoals` which have to be achieved in a pre-specified order (see e.g. (Maybury, 1991; Moore, 1994) for comparable approaches).

`fmw1_2_01: der Termin den wir neulich abgesprochen haben am zehnten an dem Samstag` (MOTIVATE)
(*the date we recently agreed upon, the 10th that Saturday*)

`da kann ich doch nich'` (REJECT)
(*then I can not*)

`wir sollten einen anderen ausmachen` (INIT)
(*we should make another one*)

`mps1_2_02: wenn ich da so meinen Termin-Kalender anschaue,` (DELIBERATE)
(*if I look at my diary*)

`das sieht schlecht aus` (REJECT).
(*that looks bad*)

Figure 4: Part of an example dialogue

Since the VERBMOBIL system is not actively participating in the appointment scheduling task but only mediating between two dialogue participants it has to be assumed that every utterance, even if it is not consistent with the dialogue model, is a legal dialogue step. The first strategy for error recovery therefore is based on the hypothesis that the attribution of a dialogue act to a given utterance has been incorrect or rather that an utterance has various facets, i.e. multiple dialogue act interpretations. Currently, only the most plausible dialogue act is provided by the semantic evaluation component. To find out whether there might be an additional interpretation the plan recognizer relies on information provided by the statistics module. If an incompatible dialogue act is encountered, an alternative dialogue act is looked up in the statistical module which is most likely to come after the preceding dialogue act and which can be consistently followed by the current dialogue act, thereby gaining an admissible dialogue act sequence.

To illustrate this principle we show a part of the processing of two turns (`fmw1_2_01` and `mps1_2_02`, see figure 4) from an example dialogue with the dialogue act assignments as provided by the semantic evaluation component. The translations stick to the German words as close as possible and are not provided by VERBMOBIL. The trace of the dialogue component is given in figure 5, starting with processing of INIT.

```
 ...
 Planner: -- Processing INIT
 Planner: -- Processing DELIBERATE
 Warning -- Repairing...
 Planner: -- Processing REJECT

Trying to find a dialogue act to bridge
DELIBERATE and REJECT ...

Possible insertions and their scores:
((SUGGEST 81326)
 (REQUEST_COMMENT 37576)
 (DELIBERATE 20572))

Testing SUGGEST for compatibility with
surrounding dialogue acts...

The previous dialogue act INIT
has an additional reading of SUGGEST:
INIT -> INIT SUGGEST !

 Warning -- Repairing...
 Planner: -- Processing INIT
 Planner: -- Processing SUGGEST
 ...
```

Figure 5: Example of statistical repair

In this example the case for statistical repair occurs when a REJECT does not – as expected – follow a SUGGEST. Instead, it comes after the INIT of the topic to be negotiated and after a DELIBERATE. The latter dialogue act can occur at any point of the dialogue; it refers to utterances which do not contribute to the negotiation as such and which can be best seen as "thinking aloud". As first option, the plan recognizer tries to repair this state using statistical information, finding a dialogue act which is able to connect INIT and REJECT[1]. As can be seen in figure 5 the dialogue acts REQUEST_COMMENT, DELIBERATE, and SUGGEST can be inserted to achieve a consistent dialogue. The annotated scores are the product of the transition probabilities times 1000 between the previous dialogue act, the potential insertion and the current dialogue act which are provided by the statistic module. Ordered according to their scores, these candidates for insertion are tested for compatibility with either the previous or the current dialogue act. The notion of compatibility refers to dialogue acts which have closely related meanings or which can be easily realized in one utterance.

[1]Because DELIBERATE has only the function of "social noise" it can be omitted from the following considerations.

To find out which dialogue acts can be combined we examined the corpus for cases where the repair mechanism proposes an additional reading. Looking at the sample dialogues we then checked which of the proposed dialogue acts could actually occur together in one utterance, thereby gaining a list of admissible dialogue act combinations. In the VERBMOBIL corpus we found that dialogue act combinations like SUGGEST and REJECT can never be attributed to one utterance, while INIT can often also be interpreted as a SUGGEST therefore getting a typical follow-up reaction of either an acceptance or a rejection. The latter case can be found in our example: INIT gets an additional reading of SUGGEST.

In cases where no statistical solution is possible plan-based repair is used. When an unexpected dialogue act occurs a plan operator is activated which distinguishes various types of repair. Depending on the type of the incoming dialogue act specialized repair operators are used. The simplest case covers dialogue acts which can appear at any point of the dialogue, as e.g. DELIBERATE and clarification dialogues (CLARIFY_QUERY and CLARIFY_ANSWER). We handle these dialogue acts by means of repair in order to make the planning process more efficient: since these dialogue acts can occur at any point in the dialogue the plan recognizer in the worst case has to test for every new utterance whether it is one of the dialogue acts which indicates a deviation. To prevent this, the occurrence of one of these dialogue acts is treated as an unforeseen event which triggers the repair operator. In figure 5, the plan recognizer issues a warning after processing the DELIBERATE dialogue act, because this act was inserted by means of a repair operator into the dialogue structure.

## 5 Conclusion

This paper presents the method for statistical dialogue act prediction currently used in the dialogue component of VERBMOBIL. It presents plan repair as one example of its use.

The analysis of the statistical method shows that the prediction algorithm shows satisfactory results when deviations from the main dialogue model are excluded. If dialogue acts for deviations are included, the prediction rate drops around 10%. The analysis of the hit rate shows also a large variation in the structure of the dialogues from the corpus. We currently integrate the speaker direction into the prediction process which results in a gain of up to 5 % in the prediction hit rate. Additionally, we investigate methods to cluster training dialogues in classes with a similar structure.

An important application of the statistical prediction is the repair mechanism of the dialogue plan recognizer. The mechanism proposed here contributes to the robustness of the whole VERBMOBIL system insofar as it is able to recognize cases where dialogue act attribution has delivered incorrect or insufficient results. This is especially important because the input given to the dialogue component is unreliable when dialogue act information is computed via the keyword spotter. Additional dialogue act readings can be proposed and the dialogue history can be changed accordingly.

Currently, the dialogue component processes more than 200 annotated dialogues from the VERBMOBIL corpus. For each of these dialogues, the plan recognizer builds a dialogue tree structure, using the method presented in section 4, even if the dialogue structure is inconsistent with the dialogue model. Therefore, our model provides robust techniques for the processing of even highly unexpected dialogue contributions.

In a next version of the system it is envisaged that the semantic evaluation component and the keyword spotter are able to attribute a *set* of dialogue acts with their respective probabilities to an utterance. Also, the plan operators will be augmented with statistical information so that the selection of the best possible follow-up dialogue acts can be retrieved by using additional information from the plan recognizer itself.